%% file: main.tex
\documentclass[sigconf]{acmart}

\AtBeginDocument{%
  \providecommand\BibTeX{{%
    \normalfont B\kern-0.5em{\scshape i\kern-0.25em b}\kern-0.8em\TeX}}}

\usepackage{booktabs} % For formal tables
\setcopyright{rightsretained}
\usepackage{url}  %Required
\usepackage{makecell}
\usepackage{graphicx}
\usepackage{caption}
\usepackage{courier}
\usepackage{multirow}
\usepackage{color}
\usepackage{subcaption}
\usepackage{color}
\usepackage{gensymb}
\usepackage{enumitem}
\usepackage{multirow}
\usepackage{adjustbox}
\usepackage[linesnumbered, ruled]{algorithm2e}
\usepackage{amsmath}

\newcommand{\nop}[1]{}

\copyrightyear{2021}
\acmYear{2021}
\setcopyright{acmcopyright}\acmConference[WSDM '21]{Proceedings of the Fourteenth ACM International Conference on Web Search and Data Mining}{March 8--12, 2021}{Virtual Event, Israel}
\acmBooktitle{Proceedings of the Fourteenth ACM International Conference on Web Search and Data Mining (WSDM '21), March 8--12, 2021, Virtual Event, Israel}
\acmPrice{15.00}
\acmDOI{10.1145/3437963.3441750}
\acmISBN{978-1-4503-8297-7/21/03}
\begin{document}

\title{Relation-aware Meta-learning for E-commerce\\ Market Segment Demand Prediction with Limited Records}
\author{Jiatu Shi{$^{1*}$}, Huaxiu Yao$^{2*}$, Xian Wu$^3$, Tong Li{$^{1}$}, Zedong Lin{$^{1}$}, Tengfei Wang{$^{1}$}, Binqiang Zhao{$^{1}$}}
\affiliation{%
  \institution{{$^{1}$}Alibaba Group, {$^{2}$}Pennsylvania State University, {$^{3}$}University of Notre Dame}
  \country{}
}
\email{{jiatu.sjt, litong.lt, zedong.lzd, wenlin.wtf, binqiang.zhao}@alibaba-inc.com}
\email{huaxiuyao@psu.edu, xwu9@nd.edu}
\thanks{*: equal contribution. Order is determined through dice rolling. Correspondence to: Xian Wu (xwu9@nd.edu)}

\input{abstract}

\maketitle
\input{introduction}
\input{relatedwork}
\input{preliminary}
\input{method}
\input{system}
\input{experiment}
\input{conclusion}
\bibliographystyle{ACM-Reference-Format}
\bibliography{ref}
\end{document}

%% file: abstract.tex
% !TEX root = main.tex
\begin{abstract}
E-commerce business is revolutionizing our shopping experiences by providing convenient and straightforward services. One of the most fundamental problems is how to balance the demand and supply in market segments to build an efficient platform. While conventional machine learning models have achieved great success on data-sufficient segments, it may fail in a large-portion of segments in E-commerce platforms, where there are not sufficient records to learn well-trained models. In this paper, we tackle this problem in the context of market segment demand prediction. The goal is to facilitate the learning process in the target segments by leveraging the learned knowledge from data-sufficient source segments. Specifically, we propose a novel algorithm, \emph{RMLDP}, to incorporate a multi-pattern fusion network (MPFN) with a meta-learning paradigm. The multi-pattern fusion network considers both local and seasonal temporal patterns for segment demand prediction. In the meta-learning paradigm, transferable knowledge is regarded as the model parameter initialization of MPFN, which are learned from diverse source segments. Furthermore, we capture the segment relations by combining data-driven segment representation and segment knowledge graph representation and tailor the segment-specific relations to customize transferable model parameter initialization. Thus, even with limited data, the target segment can quickly find the most relevant transferred knowledge and adapt to the optimal parameters. We conduct extensive experiments on two large-scale industrial datasets. The results justify that our RMLDP outperforms a set of state-of-the-art baselines. Besides, RMLDP has been deployed in Taobao, a real-world E-commerce platform. The online A/B testing results further demonstrate the practicality of RMLDP.
\end{abstract}

%% file: introduction.tex
% !TEX root = main.tex
% 1. 这个问题很重要，对业务非常重要， 两个方面：1. 对用户的意义，对公司的意义
%当下，国内外的大规模电商零售平台（如，亚马逊，淘宝）为我们的日常消费提供了极大地便利。为了建设一个更加高效的平台，
%对平台的市场进行细化并了解细分市场的需求成为了一个最基础的问题。
%细分市场的需求预估越精准，我们就能更好地指导商家生产，备货，减少库存积压，并更好地处理消费者流量资源的分配。
%受整体供应链上下游滞后的影响，单纯的实时的需求预测（例如预估明天的消费需求）就显得力不从心，作为补充，
%我们重点关注基于历史数据进行提前若干周甚至月对未来的一个时间段进行长周期需求预估。

% 2. 简单的解决方法是时间序列的预测 -> 但是这不能用来做小样本， 小样本的意义

%在对细分市场长周期的需求预估中，季节性和趋势知识广泛存在于不同行业中。比如，
%随着夏天的临近，家电市场空调的销量会攀升，而取暖器的需求会骤减。天气转凉时，服装市场T恤的需求会下降，而羽绒服的需求会上升。
%传统的回归预测模型如GBDT，Random Forest能够较好地在样本充足的情况下拟合这些知识。而对于那些样本有限的类目，拟合就比较困难。
%对于样本有限的细分市场来说，由于缺乏业务效果数据，难以获得展示机会，从而难以获得新数据，形成恶性循环，
%大量的长尾类目得不到扶持，缺乏上升渠道，最终体现为整个平台同质化严重。
%而小样本类目不等同于优质类目，单纯依托运营人为经验扶持，会伤害了热门类目的产出，影响整体的分配效率。
%所以构建更为精准的预估模型迫在眉睫。

% 3. 为了做小样本，存在以下两个挑战：（1）公司层面上end-to-end，模型灵活性，对应maml （2）不同类目之间的关系；对应task-embedding和knowledge graph
%因此，我们提出了xxx方法来解决这个问题
%贡献：问题，方法，实验
\section{Introduction}
\label{sec:intro}
\begin{figure}[h]
	\centering
 	\includegraphics[width=0.38\textwidth]{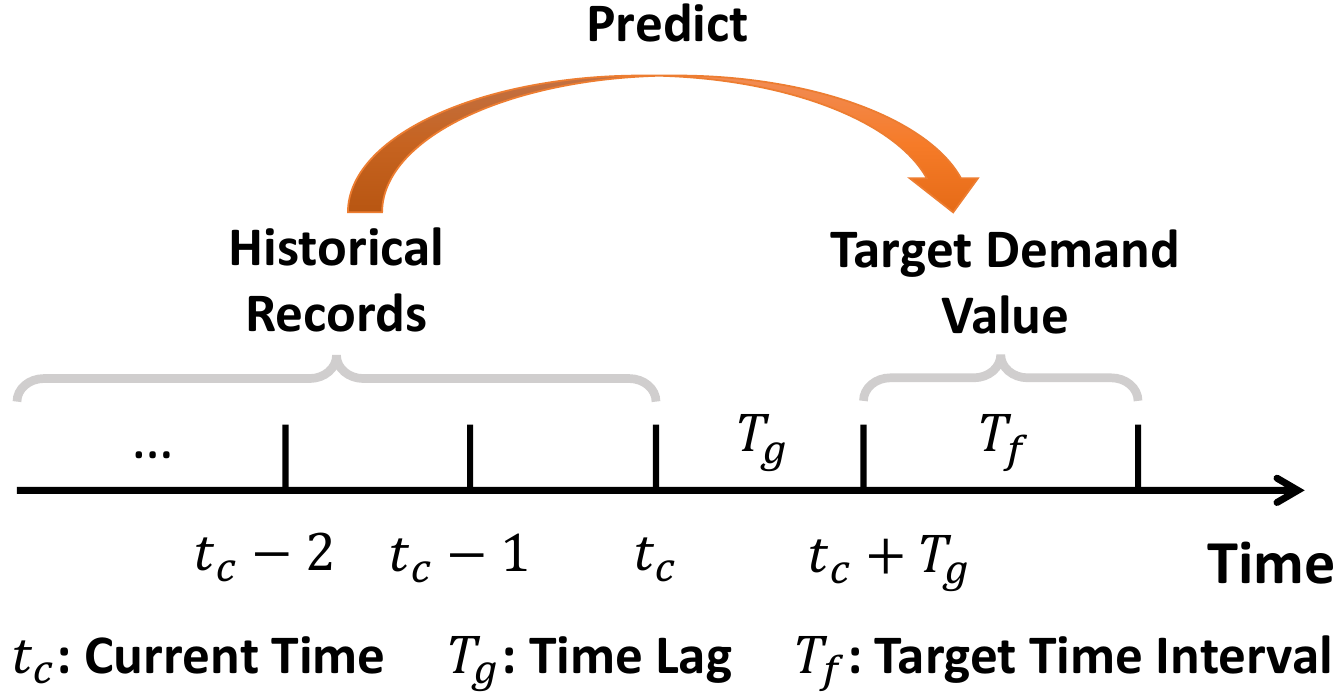}
	\caption{Illustration of Segment Demand Prediction.} 
	\label{fig:prediction_illustration}
\end{figure}

Large-scale E-commerce platforms (e.g., Amazon, Taobao) are revolutionizing people's shopping experience by providing numerous merchandise options at one's fingertips. To build an efficient E-commerce platform, one of the most fundamental problems is how to balance the demand and supply in the market, which requires an accurate demand prediction model for every market segment (e.g., wallet, belt). An accurate demand prediction model benefits the platform from three aspects: 1) pre-allocate resources to meet the market demand; 2) reduce the backlog of commodities; 3) optimize the allocation strategies of traffic source. In addition, due to the lags between upstream and downstream of the supply chain, real-time segment demand prediction (e.g., predict the next day's demand) may be impractical. Instead, given the historical demand records, we study the demand prediction problem as predicting the demand value of a future target period (e.g., one month) several weeks in advance (as illustrated in Figure~\ref{fig:prediction_illustration}).
%In the long-term market segment demand prediction, seasonal trend is a common property across different market segments. For example, with the approaching of summer, the sales volume of air conditioners in home appliance market will increase, and the demand for heaters will sharply decrease. When the weather turns cold, the demand for T-shirts in the clothing market will decrease, while the demand for down coats will increase. 

\begin{figure}[h]
	\centering
 	\includegraphics[width=0.4\textwidth]{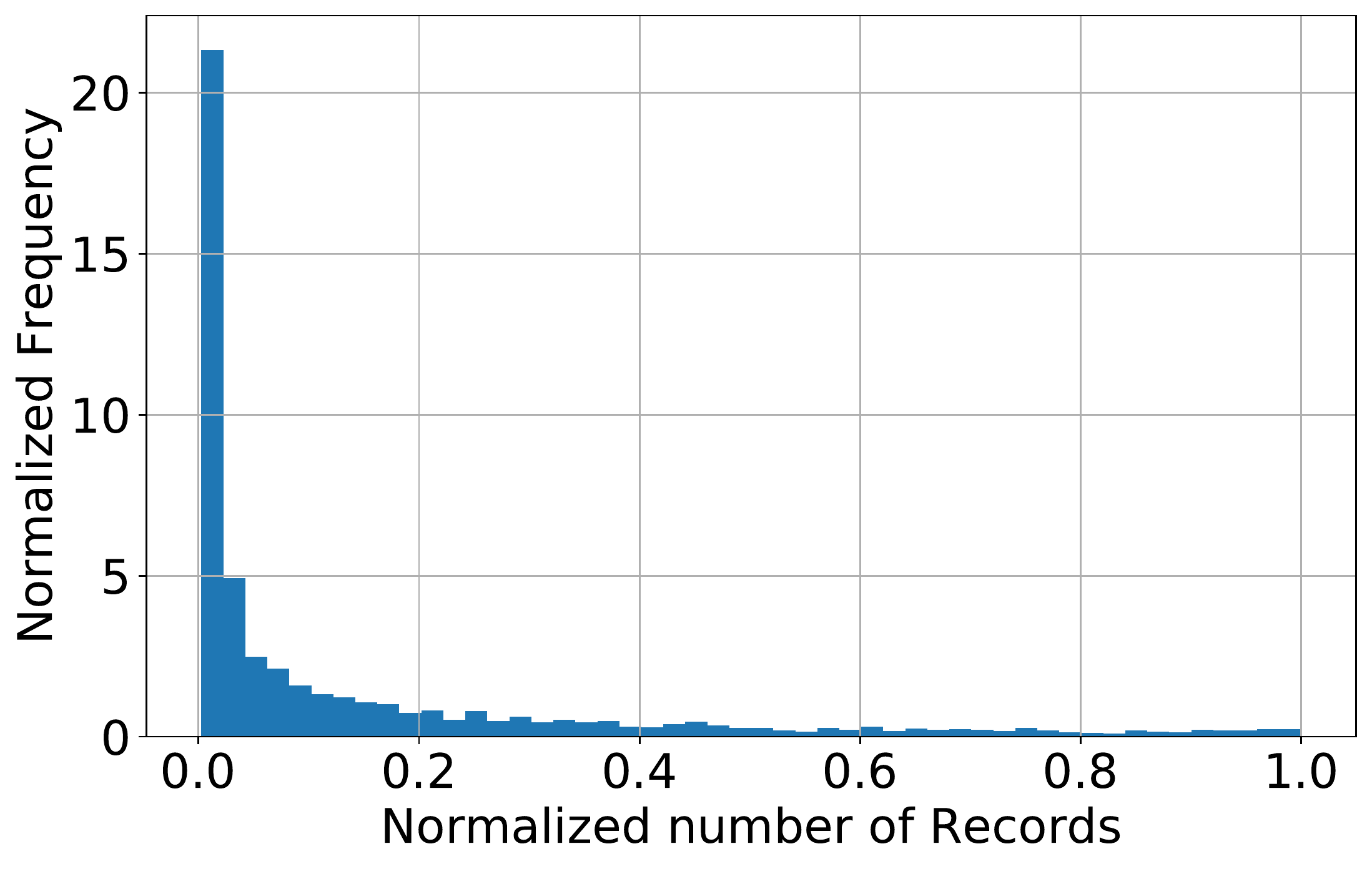}
	\caption{Histogram of segment records' frequencies in Juhuasuan. Both frequency and number of records are normalized due to privacy policy. A large number of segments only have limited records.} 
	\label{fig:frequency}
\end{figure}

To predict market segment demand, traditional ensemble models (e.g., XGBoost~\cite{chen2016xgboost}) and advanced deep learning methods (e.g., LSTM~\cite{hochreiter1997long}, GRU~\cite{chung2014empirical}) are capable of capturing time-varying sequential patterns (e.g., seasonal trend) and making accurate predictions. The superiority of these methods relies on large-scale labeled training data. Unfortunately, as illustrated in Figure~\ref{fig:frequency}, a large portion of market segments is located in a long-tail position with limited records, which leads to poor prediction performance and, in turn, affects the efficiency of the platform. The reasons are two-fold: 1) Usually, market segments with more records are more likely to be exhibited on the platform. For data-insufficient market segments, the lack of exposure opportunities brings difficulties in collecting new records and negatively affects the model performance on these segments. The process finally forms a vicious circle, resulting in the platform's homogenization; 2) The data-insufficient market segments have data quality issues. Due to the limited resources (e.g., the number of exhibitions), for data-insufficient market segments, purely rely on platform managers' support is impractical and may jeopardize the performance of the mainstream segments. Therefore, how to improve the prediction performance for market segments with limited data remains a non-trivial but necessary problem.

%Traditional regression prediction models such as GBDT~\cite{} and random forest~\cite{} are able to capture these knowledge with sufficient records. 
%But for those segments with limited records, fitting is more difficult. 

To tackle this ``small data" problem, recently, knowledge transfer (e.g., transfer learning, meta-learning)~\cite{pan2009survey,finn2017model} has achieved great success in a series of applications, such as computer vision~\cite{tzeng2017adversarial}, natural language processing~\cite{li2019transferable}. To improve target tasks' learning process with limited labeled data, knowledge transfer leverages the prior knowledge learned from relevant source tasks. In the segment demand prediction, only applying conventional knowledge transfer algorithms to improve the performance of data-insufficient segments faces the following two major challenges:
\begin{itemize}[leftmargin=*]
    \item \textbf{C1: How to boost the stability and generalization ability of knowledge transfer?} Usually, the performance of knowledge transfer relies on the similarity of distributions between source and target tasks. Significant data distribution difference between tasks may lead to unstable transfer or even worse prediction performance. Therefore, a sufficiently generalized knowledge transfer framework is required, covering comprehensive and diverse temporal patterns of market segments. 
    \item \textbf{C2: How to incorporate the complex relations among market segments?} It is non-trivial to capture the complex segment relations using traditional knowledge transfer methods (e.g., fine-tuning), where the transferable knowledge is globally shared across source segments. However, in E-commerce platforms, the differences between segments can not be overlooked, and thus the globally shared transferable knowledge may not be robust enough to all scenarios. For example, the demands of down jackets are probably similar to coats' demands, while dissimilar to t-shirts. Thus, segment relations are necessary to be incorporated in knowledge transfer framework.
\end{itemize}

Hence, to address the above challenges, in this paper, we propose a novel framework \textbf{RMLDP} for data-insufficient market segment demand prediction. The goal for RMLDP is to build a customized meta-learning paradigm upon a market demand prediction model. Specifically, we first construct a multi-pattern fusion network (MPFN) for market segment demand prediction, which jointly captures both local and seasonal temporal patterns by two Gated recurrent units (GRUs). Regrading the MPFN as the base model, the first challenge is solved by learning and transferring the model parameter initialization of the MPFN under the meta-learning paradigm. Here, various source segments sampled from diverse categories (e.g., food, clothing) are used for initialization learning. Finally, we introduce a data-driven segment representation and a segment knowledge graph representation to capture the complex segment relations. For each segment, the relational information are further used to modulate the model parameter initialization. 

In summary, our major contributions are three-fold:
\begin{itemize}[leftmargin=*]
    \item To the best of our knowledge, we are the first to study the problem of market segment demand prediction with limited data by transferring the knowledge from mainstream segments.
    \item We develop a novel framework, RMLDP, to solve the market segment demand prediction task. RMLDP incorporates a multi-pattern fusion network with the meta-learning paradigm. The segment relations are further distilled to customize the model parameter initialization in meta-learning paradigm. Furthermore, we deploy the proposed method into the online platform.
    \item We collect the market demand records from two large-scale E-commerce platforms: Juhuasuan and Tiantiantemai. Comparing with baseline methods, the superior performance of RMLDP demonstrates the effectiveness of our framework under both offline and online scenarios.
\end{itemize}

%The rest of the paper is organized as follows: we discuss the related work in Section~\ref{sec:related}. Then, some notations and concepts with a formal problem definition are given in Section~\ref{sec:preliminary}. The detailed description of our proposed RMLDP is introduced in Section~\ref{sec:model}. The deployed system is described in Section~\ref{sec:system}. In section~\ref{sec:experiments}, we evaluate our method on two large-scale datasets. Finally, we conclude our paper in Section~\ref{sec:conclusion}.

%% file: relatedwork.tex
% !TEX root = main.tex
\section{Related Work}
This section briefly discusses two categories of related work: time series prediction and knowledge transfer.\vspace{0.4em}
\label{sec:related}

\noindent \textbf{Time Series Prediction.}
Traditional approaches (e.g., ARIMA~\cite{pan2012utilizing}, Kalman filtering~\cite{lippi2013short}) have been widely used in time series applications. These methods fail to capture complex non-linear temporal correlations due to the limited expressive capability. With stronger expressive power, deep learning methods, especially recurrent neural network-based approaches (e.g., GRU~\cite{chung2014empirical} and LSTM~\cite{hochreiter1997long}), have achieved great success in time series modeling~\cite{qin2017dual,laptev2017time,lv2018lc,lai2018modeling,salinas2019deepar,wang2019deep,rangapuram2018deep}. To further improve the prediction performance, recently, more information have been incorporated in the basic recurrent neural network structures by applying attention mechanism~\cite{qin2017dual,ma2017dipole} or multi-resolution modeling~\cite{wu2018restful,huang2019online}. \emph{However, all these methods rely on large-scale training data. In contrast, our work aims to improve the prediction of data-insufficient target segments by knowledge transfer. Besides, those methods focus on the prediction for the next step/a few steps. In this work, we focus on the early forecast for a future time interval under the real-world E-commerce scenario.}\vspace{0.4em}

\noindent \textbf{Knowledge Transfer.}
To benefit the learning process on task with limited data, transferring knowledge from its related tasks has achieved great success in recent years~\cite{pan2009survey}. Conventional transfer learning methods learn transferable latent factors between one source domain and one target domain. The latent factors are captured by a series of techniques, such as matrix factorization~\cite{long2013transfer}, manifold learning~\cite{gong2012geodesic} and deep learning~\cite{tzeng2017adversarial,long2015learning}. Recently, meta-learning (a.k.a., learning to learn) provides a more stable and flexible way for knowledge transfer. The goal for meta-learning is to generalize the knowledge from various tasks and then adapt them to unseen tasks. In meta-learning, the transferable knowledge are regarded as model parameter initializations~\cite{finn2017model,finn2018probabilistic,lee2018gradient,yao2019hierarchically,finn2017meta}, metric mapping function~\cite{snell2017prototypical,vinyals2016matching,oreshkin2018tadam}, or meta-optimizer~\cite{ravi2016optimization,andrychowicz2016learning}, etc.
%In this way, the amount of data required to train these new tasks is reduced. 
In the time series related problems,~\citeauthor{oreshkin2020n} briefly discusses the relation between the neural time series prediction and meta-learning meta-learning~\cite{oreshkin2020n}. ~\citeauthor{yao2019learning} incorporates the gradient-based meta-learning with a region functionality based memory~\cite{yao2019learning} for spatiotemporal prediction. \emph{However, this method relies on the spatial semantic correlations between tasks, which limits its applicability in our problem. To the best of our knowledge, we are the first to study market segment demand prediction with limited records by borrowing relation-aware knowledge from other segments.}

%% file: preliminary.tex
\section{Preliminaries}
\label{sec:preliminary}
In this section, we define some concepts and notations and then formally define our problem. Assuming the whole market is split into \begin{small}$I$\end{small} market segments \begin{small}$\{s_1,\ldots,s_I\}$\end{small}, each market segment \begin{small}$s_i$\end{small} represents one category of products (e.g., sweaters, orange juice).

\noindent \textbf{Definition 1 (Market Demand Value)} \emph{For each segment \begin{small}$s_i$\end{small} at time step $t_i$, the market demand value \begin{small}$x_{i,t_i}$\end{small} is defined as the number of purchasing requests in a fixed time window \begin{small}$[t_i, t_i+t^{'}]$\end{small}. In this paper, the length fixed time interval \begin{small}$t^{'}$\end{small} is defined as one day (i.e., \begin{small}$t^{'}=1$\end{small}).}

\noindent \textbf{Definition 2 (Target Demand Value)} \emph{As illustrated in Figure~\ref{fig:prediction_illustration}, we aim to predict the market demand for a future target time interval \begin{small}$T_f$\end{small} several weeks in advance. Supposing the current time stamp is \begin{small}$t_c$\end{small} and the time lag between the current time and the further target time is \begin{small}$T_g$\end{small}, we define the target demand value \begin{small}$y_{i, t_c}$\end{small} as the total market demand value of \begin{small}$s_i$\end{small} between time interval \begin{small}$[t_c+T_g, t_c+T_g+T_f]$\end{small} (i.e., \begin{small}$y_{i,t_c}=\sum_{j=T_g}^{T_g+T_f} x_{i,t_c+j}$\end{small}).}

\noindent \textbf{Problem: Market Segment Demand Prediction with Limited Records}
Assuming that we have a set of diverse source segments \begin{small}$\{s_1,\ldots, s_I\}$\end{small} and a target segment \begin{small}$s_t$\end{small} with limited records, we aim to predict the target demand value \begin{small}$y_{t,t_c}$\end{small} in the testing dataset of the target segment. Additionally, for each segment \begin{small}$s_i$\end{small} at time stamp \begin{small}$t_c$\end{small}, we further introduce several statistical features \begin{small}$\mathbf{e}_{i,t_c}$\end{small} (e.g., \# of items, sellers, brands) and customers' action features (e.g., click, collect, add to cart and take order). 
%We further count up the number of four types of customers' actions  in 8 different historical time windows (1 day, 3days, 7days, 14days, 30days, 90days, 180days, 365days). 

We denote the concatenation of market demand value \begin{small}$x_{i,t_c}\in \mathbb{R}^1$\end{small} and external features \begin{small}$\mathbf{e}_{i,t_c}\in \mathbb{R}^{e-1}$\end{small} as \begin{small}$\mathbf{x}_{i,t_c}=x_{i,t_c}\oplus \mathbf{e}_{i,t_c}\in \mathbb{R}^e$\end{small}. The market segment demand prediction model (a.k.a., base model) is defined as \begin{small}$f$\end{small} with the learnable parameters \begin{small}$\theta$\end{small}. Formally, our problem is formulated as: 
\begin{equation}
\small
    y_{t,t_c}^{*}=\arg\max_{y_{t,t_c}} p(y_{t,t_c}|\theta_{0t}^{*},\{\mathbf{x}_{t,1},\ldots,\mathbf{x}_{t,t_c}\})
\end{equation}
where \begin{small}$\theta_{0t}^{*}$\end{small} denotes the segment-specific initializations, which are transferred from all source segments using the target segment information. Detailed discussions about customized model initializations are in Section~\ref{sec:transfer} and~\ref{sec:customization}. We name the process of learning transferable knowledge from source segments as \emph{meta-training} and the adaption in target segments as \emph{meta-testing}. 

% we spatially divide a %whole 
% city $c$ %$\mathcal{C}$ 
% into an $I_{c}\times J_{c}$ %$I_\mathcal{C}\times J_\mathcal{C}$
% grid map which contains $I_{c}$ rows and $J_{c}$ columns. 
% We treat each grid as a region $r_{c}$, and define the set of all regions as $\mathcal{R}_c$. 

%% file: method.tex
\section{Methodology}
\label{sec:model}
\begin{figure*}[h]
	\centering
 	\includegraphics[width=0.85\textwidth]{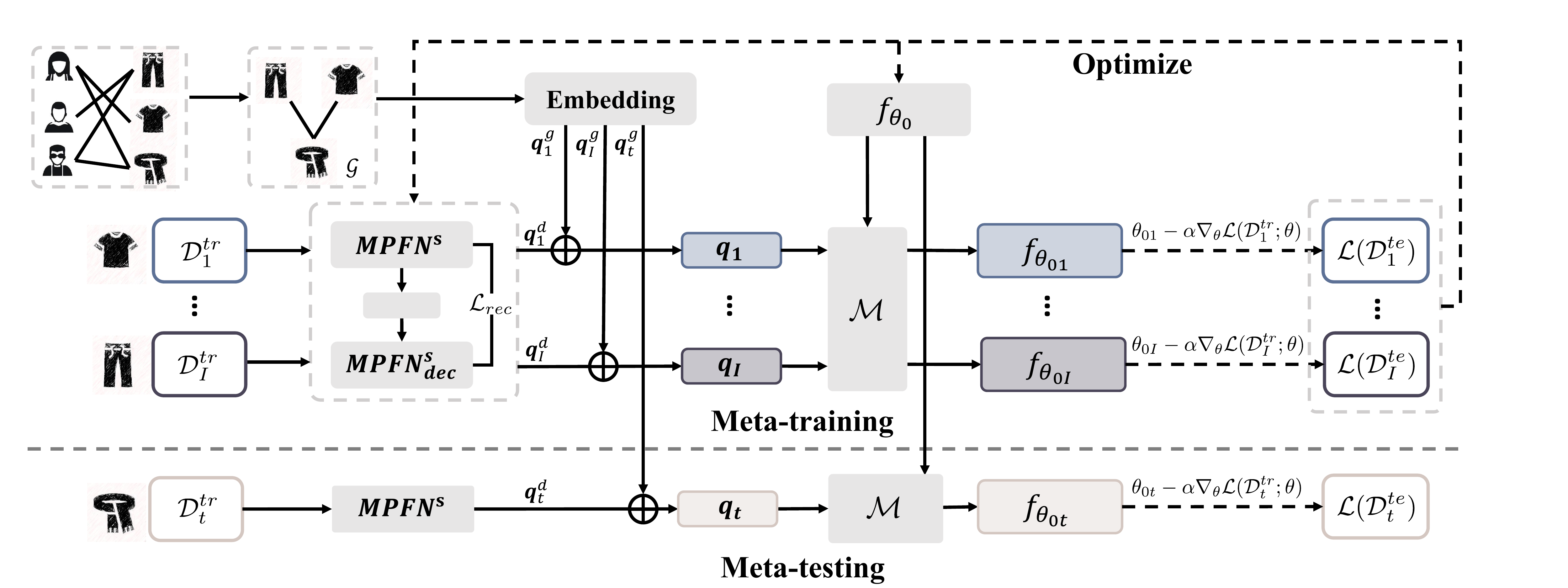}
	\caption{The Framework of RMLDP. In the meta-training process, the initialization $\theta_0$ are customized by using the data-driven segment representation (learned by $\mathrm{MPFN}^s$) and the representation of segment knowledge graph $\mathcal{G}$ (constructed by users' purchasing records). The performance on the testing sets of all source segments are further used to update all transferable knowledge (grey blocks). In the meta-testing process, the transferred knowledge is adapted to the target segment $s_t$.} 
	\label{fig:framework}
\end{figure*}
In this section, we introduce our proposed framework: RMLDP (Relation-aware Meta-Learning for Demand Prediction). The whole framework is shown in Figure~\ref{fig:framework}. The goal for RMLDP is to facilitate the learning process of data-insufficient target segment demand prediction by adapting the transferred knowledge (i.e., all grey blocks in Figure~\ref{fig:framework}) from data-sufficient source segments. In particular, the base model \begin{small}$f$\end{small} is first designed as a multi-pattern fusion network (MPFN), where both local and seasonal temporal patterns are considered. Then, RMLDP incorporates the base model \begin{small}$f$\end{small} and the meta-learning paradigm, where the model parameter initialization are regarded as transferable knowledge. To further modulate the parameter initialization, we distill knowledge from segment representations, including a data-driven segment representation and a segment knowledge graph representation. We detail three key components in the following subsections: \emph{multi-pattern fusion network, knowledge transfer, and adaptation, relation-aware modulation}.
\subsection{Multi-pattern Fusion Network}
\begin{figure}[h]
	\centering
 	\includegraphics[width=0.42\textwidth]{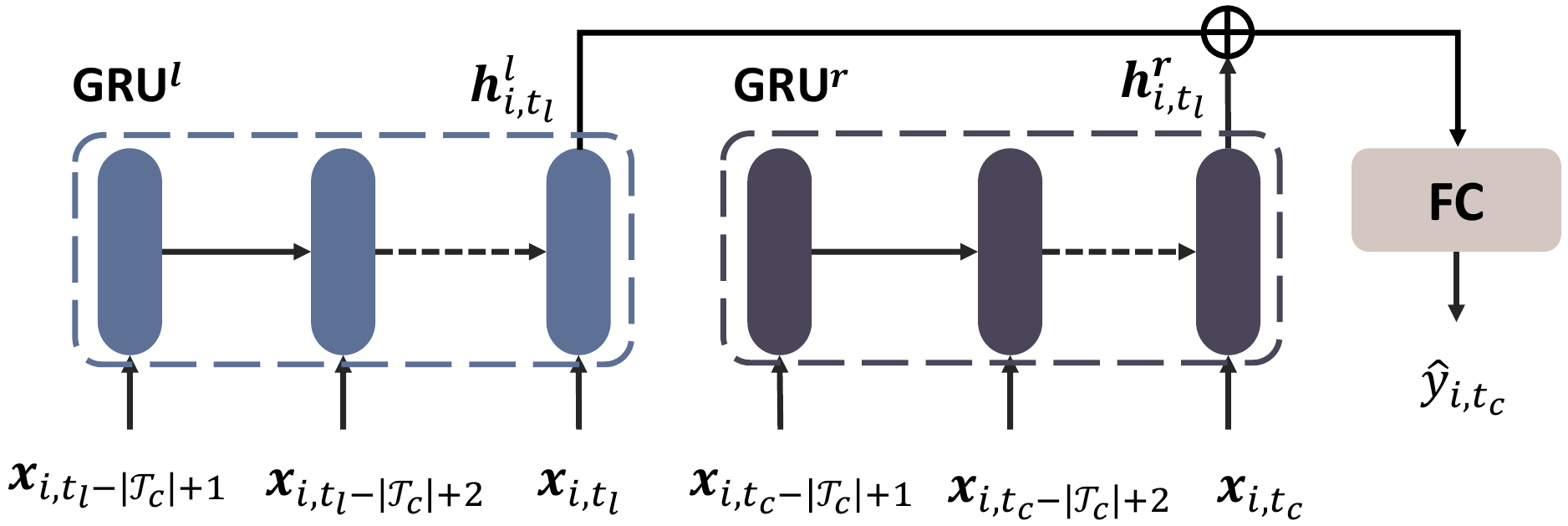}
	\caption{Illustration of MPFN for market segment demand prediction. $\mathrm{GRU}^l$ and $\mathrm{GRU}^r$ capture seasonal and local temporal patterns, respectively.} 
	\label{fig:mpfn}
\end{figure}

In this subsection, we propose a multi-pattern fusion network (\textbf{MPFN}) for market segment demand prediction. The framework is illustrated in Figure~\ref{fig:mpfn}. The goal for MPFN is to predict the target demand value by capturing the temporal patterns from the historical records. To achieve this goal, we adopt a GRU network to capture non-linear relations among historical records. Concretely, for predicting the target demand value \begin{small}$y_{i, t_c}$\end{small} of segment \begin{small}$s_i$\end{small}, the most recent \begin{small}$|\mathcal{T}_c|$\end{small} demand values (i.e., \begin{small}$\{\mathbf{x}_{i,t_c-|\mathcal{T}_c|+1},\ldots,\mathbf{x}_{i,t_c}\}$\end{small}) are fed into the GRU, which is formulated as:
\begin{equation}
\small
\mathbf{h}^r_{i,t_c}=\mathrm{GRU}^r(\mathbf{h}^r_{i,t_c-1};\mathbf{x}_{i,t_c}).
\end{equation}
The temporal representation \begin{small}$\mathbf{h}^r_{i,t_c}$\end{small} encodes the local temporal patterns from the closest records.

As mentioned in Section~\ref{sec:preliminary}, different from real-time demand prediction, there exists a time lag \begin{small}$T_g$\end{small} between current time and the target prediction time. Thus, the temporal patterns captured from closest demand records are probably insufficient to achieve satisfactory performance. Fortunately, seasonal temporal patterns provide us with useful periodic information. For example, the demand trend for winter coat in this December is similar to the trend in the last December. However, as suggested in~\cite{yao2018modeling}, it is non-trivial to train a single GRU network for handling long-term seasonal patterns due to the risk of gradient vanishing. Instead, another GRU network is introduced to model the seasonal patterns as:
\begin{equation}
\small
    \mathbf{h}^l_{i,t_l}=\mathrm{GRU}^l(\mathbf{h}^l_{i,t_l-1};\mathbf{x}_{i,t_l}),
\end{equation}
where \begin{small}$t_l=t_c+T_g-365$\end{small} represents the corresponding historical time of the target demand value (i.e., same day in the last year). The sequence \begin{small}$\{\mathbf{x}_{i,t_l-|\mathcal{T}_c|+1}\ldots \mathbf{x}_{i,t_l}\}$\end{small} are fed into \begin{small}$\mathrm{GRU}^l$\end{small}. 

By fusing the hidden representations \begin{small}$\mathbf{h}^l_{i,t_l}$\end{small} and \begin{small}$\mathbf{h}^c_{i,t_c}$\end{small} as \begin{small}$\mathbf{\hat{h}}_{i,t_c}=\mathbf{h}^l_{i,t_l}\oplus \mathbf{h}^c_{i,t_c}$\end{small}, both local and seasonal temporal patterns are captured. Then, we use one fully connected layer for prediction as:
\begin{equation}
\small
    \hat{y}_{i,t_c}=\mathbf{W}_f\mathbf{\hat{h}}_{i,t_c}+\mathbf{b}_f,
\end{equation}
where \begin{small}$\mathbf{W}_{f}$\end{small} and \begin{small}$\mathbf{b}_f$\end{small} are learnable parameters. In this paper, mean square error (MSE) is used as loss function as:
\begin{equation}
\small
\label{eqn:loss}
    \mathcal{L}=\sum_{t_c}(y_{i,t_c}-\hat{y}_{i,t_c})^2.
\end{equation}
As mentioned in Section~\ref{sec:preliminary}, the MPFN is regarded as base model \begin{small}$f$\end{small} with all learnable parameters are denoted as \begin{small}$\theta$\end{small}.

% \end{equation}
\subsection{Knowledge Transfer and Adaptation}
\label{sec:transfer}
After constructing the base model MPFN, we discuss the meta-learning paradigm, which transfers knowledge from source segments to the target segment with limited data. To increase knowledge transfer stability, the transferable knowledge is expected to be general enough and contain comprehensive relations between market segments and their historical temporal patterns. %for adapting the knowledge to an unseen segment under different scenarios.
% As we discussed before, to increase the stability of knowledge transfer, we transfer the knowledge learned from multiple segments. 

Motivated by the model-agnostic meta-learning (MAML)~\cite{finn2017model}, these transferable knowledge are encrypted in the model parameter initialization \begin{small}$\theta_0$\end{small} of base model \begin{small}$f$\end{small}. Thus, the aim of knowledge transfer is to learn an optimal model parameter initialization from multiple source segments \begin{small}$\{s_1, \ldots, s_{I}\}$\end{small}. For segment \begin{small}$s_i$\end{small}, the model parameters \begin{small}$\theta$\end{small} of market demand prediction are updated starting from \begin{small}$\theta_0$\end{small} as follows:
\begin{equation}
\small
\theta_i=\theta_0-\alpha\nabla_\theta \mathcal{L}(\mathcal{D}_{i}^{tr};\theta),
\label{eqn:maml_inner}
\end{equation}
where the empirical risk \begin{small}$\mathcal{L}$\end{small} is defined as mean square error in Eqn.~\eqref{eqn:loss}. \begin{small}$\mathcal{D}_i^{tr}=\{\mathbf{X}^{tr}_{i,t_c},y^{tr}_{i,t_c}\}_{t_c=1}^{N^{tr}}$\end{small} is the training set sampled from segment \begin{small}$s_i$\end{small}, where \begin{small}$N^{tr}$\end{small} denotes the number of training samples and \begin{small}$\mathbf{X}^{tr}_{i,t_c}=\{\mathbf{x}^{tr}_{i,t_l-|\mathcal{T}_c|+1},\ldots, \mathbf{x}^{tr}_{i,t_l}, \mathbf{x}^{tr}_{i,t_c-|\mathcal{T}_c|+1},\ldots, \mathbf{x}^{tr}_{i,t_c}\}$\end{small} represents the used demand sequence in MPFN. 
%In Eqn.~\eqref{eqn:maml_inner}, we use one exemplar gradient step but usually more gradient steps can be used in practice. 

After getting the segment-specific parameter \begin{small}$\theta_i$\end{small}, we sample the testing dataset \begin{small}$\mathcal{D}_i^{te}=\{\mathbf{X}^{te}_{i,t_c},y_{i,t_c}^{te}\}_{t_c=1}^{N^{te}}$\end{small} from \begin{small}$s_i$\end{small} to update the model parameter initialization \begin{small}$\theta_0$\end{small} by minimizing the empirical risk as:
\begin{equation}
\small
\label{eq:meta_loss}
\theta_0\leftarrow \min_{\theta_0}\sum_{i=1}^{|I|}\mathcal{L}(\mathcal{D}_i^{te};\theta_i)
\end{equation}
where \begin{small}$|I|$\end{small} denotes the number of source segments. At the end of meta-training process, we get \begin{small}$\theta^{*}_0$\end{small} as the learned optimal model parameter initialization. 
%Thus, the learned optimal initialization $\theta^{*}_0$ encrypts the relation between symptoms and diseases.

Given an target segment \begin{small}$s_t$\end{small}, the segment-specific parameter \begin{small}$\theta_t$\end{small} is achieve by performing gradient descent starting from the learned initialization \begin{small}$\theta_0^{*}$\end{small} with the training data \begin{small}$\mathcal{D}_t^{tr}$\end{small}, i.e., 
\begin{equation}
\small
\theta_t=\theta_0^{*}-\alpha\nabla_{\theta}\mathcal{L}(\mathcal{D}_t^{tr};\theta).
\label{eq:maml_new}
\end{equation}
We finally evaluate the performance by the testing set \begin{small}$\mathcal{D}_{t}^{te}$\end{small} of segment \begin{small}$s_t$\end{small} using adapted parameter \begin{small}$\theta_t$\end{small}.
\subsection{Relation-aware Customization}
\label{sec:customization}
The above knowledge transfer and adaptation framework regards the transferable knowledge as the globally shared model parameter initialization \begin{small}$\theta_0^{*}$\end{small} across all source segments. However, the globally shared knowledge may incapable of well-capturing underlying complex segment relations. For example, supposing we need to predict Men's clothing's market demand, both local and seasonal temporal trends are similar to clothing from other groups (e.g., women, children), while the temporal trends are probably dissimilar to the electric appliances. 
%A shared transferable knowledge (i.e., model parameter initialization)  these segment-specific relations.
Thus, in this section, we tailor the segment-specific relations to modulate the model parameter initialization. Specifically, we consider two types of segment relational representations: data-driven segment representation and segment knowledge graph representation. The data-driven segment representation implicitly encrypts the segment relations. Generated by users' purchase records, the segment knowledge graph further explicitly models the relations among different segments. We detail these two types of representations in the following subsections.
\subsubsection{Data-driven Segment Representation}
For data-driven segment representation, we encode the segment-specific information into one representation vector. The relations among segments are implicitly included in the representations since similar segments have similar representations. As suggested in~\cite{vuorio2018toward,yao2019hierarchically}, learning the representation of each segment $s_i$ is equal to aggregate the training data \begin{small}$\mathcal{D}_{i}^{tr}$\end{small} to a representation vector. Here, we introduce one MPFN as aggregator denoted as \begin{small}$\mathrm{MPFN}^{s}$\end{small}. The aggregator first encodes each data sample into one vector and then a sample-level mean pooling layer is applied on the top of encoder. Formally, the aggregation process is formulated as:
\begin{equation}
\small
\label{eq:segment_repr}
    \mathbf{q}_{i}^d=\frac{1}{N^{tr}}\sum_{t_c}\mathrm{MPFN}^s(\mathbf{X}_{i,t_c}^{tr}).
\end{equation}
%\yao{The data-driven market segment representation depends on the and encode }
Empirically, only using the loss signal defined in Eqn.~\eqref{eq:meta_loss} to guide the segment representation learning is difficult. To increase the stability of segment representation learning, we introduce the reconstruction loss with a decoder \begin{small}$\mathrm{MPFN}^{s}_{dec}$\end{small}, which is defined as:
\begin{equation}
\small
\label{eq:rec}
    \mathcal{L}_{rec}=\frac{1}{N^{tr}}\sum_{t_c}\Vert \mathbf{X}_{i,t_c}^{tr} - \mathrm{MPFN}^s_{dec}(\mathrm{MPFN}^s(\mathbf{X}_{i,t_c}^{tr})) \Vert_F^2
\end{equation}
where \begin{small}$\Vert\cdot \Vert_F$\end{small} is defined as Frobenius norm.

\subsubsection{Segment Knowledge Graph Representation}
In real-world E-commerce platforms, the relations between segments can further be reflected by users' purchasing records. For each pair of segments, their similarity is proportional to the frequency of co-occurrence in the same order. For example, women usually purchase sweater and skirt together. But it is unlikely to purchase shampoo and refrigerator at the same time. Given users' purchasing records, we build a segment knowledge graph \begin{small}$\mathcal{G}$\end{small}, where each node \begin{small}$n_i$\end{small} in the knowledge graph represents one segment. For each pair of nodes \begin{small}$n_u$\end{small} and \begin{small}$n_v$\end{small}, the link weight \begin{small}$\omega_{uv}$\end{small} is calculated by the co-occurrence frequency in the same order. We further set a threshold to filter some low similarity links.

Then, to map each segment into a fixed low dimensional space and maintain their relational structure, we adopt Deepwalk~\cite{perozzi2014deepwalk} on the constructed knowledge graph \begin{small}$\mathcal{G}$\end{small}. The ad-hoc learned graph embedding vector $\mathbf{u}^g_{i}$ is feed into a graph convolutional layer to get the representation of each segment in the knowledge graph, which is denoted as \begin{small}$\mathbf{q}^g_{i}=\mathrm{FC}_{\mathbf{W}_{emb}}(\mathbf{u}^g_{i})$\end{small}. Note that the data-driven segment representation and the segment knowledge graph representation are mutually complementary. In data-driven segment representation, the similarity of segments mainly reflects their temporal patterns. By contrast, the similarity of segments in this knowledge graph are guided by users' purchasing records. 

\subsubsection{Relation Fusion and Knowledge Modulation}
\label{sec:representation}
After generating the data-driven segment representation \begin{small}$\mathbf{q}^d_{i}$\end{small} and the knowledge graph representation \begin{small}$\mathbf{q}_{i}^{g}$\end{small}, we then fuse these two types of representations and get the final segment-specific representation as: \begin{small}$\mathbf{q}_{i}=\mathbf{q}_{i}^d\oplus \mathbf{q}_i^{g}$\end{small}. To customize the globally shared model parameter initialization \begin{small}$\theta_0$\end{small}, we introduce a modulating function \begin{small}$\mathcal{M}(\cdot)$\end{small}, which consists of a mapping layer with an activation function. The modulating function is defined as:
\begin{equation}
\small
    \mathcal{M}(\mathbf{q}_{i})=\sigma(\mathbf{W}_m\mathbf{q}_{i}+\mathbf{b}_m),
\end{equation}
where \begin{small}$\mathbf{W}_m$\end{small} and \begin{small}$\mathbf{b}_m$\end{small} are trainable parameters. By using the modulating function, the segment representation is mapped to the same space of the model parameter initialization \begin{small}$\theta_0$\end{small}. Then, the customization process is formulated as:
\begin{equation}
\small
    \theta_{0i}=\mathcal{M}(\mathbf{q}_{i})\odot \theta_0,
\label{eq:customize}
\end{equation}
Here \begin{small}$\theta_{0i}$\end{small} represents the task specific parameter initialization. Then, for segment \begin{small}$i$\end{small}, we perform the gradient steps starting from the customized initialization \begin{small}$\theta_{0i}$\end{small} rather than \begin{small}$\theta_0$\end{small}.

By combining the empirical risk \begin{small}$\mathcal{L}$\end{small} in Eqn.~\eqref{eq:meta_loss} and the reconstruction loss \begin{small}$\mathcal{L}_{rec}$\end{small} in Eqn.~\eqref{eq:rec}, we revise the objective function in Eqn.~\eqref{eq:meta_loss} and formulate the final objective function as:
\begin{equation}
\label{eq:joint_loss}
\small
    \min_{\Theta}\mathcal{L}_{joint}=\min_{\Theta}\sum_{i=1}^{|I|}\mathcal{L}(\mathcal{D}_i^{te};\theta_i)+\lambda \mathcal{L}_{rec},
\end{equation}
where the hyperparameter $\lambda$ is used to balance the value of two loss terms. We describe learning process for RMLDP in Algorithm~\ref{alg:RMLDP}.
%For the meta-testing process on target segments, we detail it in Appendix~\ref{sec:app_metatesting}.
\begin{algorithm}[tb]
    \caption{Meta-training Framework for RMLDP}
    \label{alg:RMLDP}
    \KwIn{source segments \begin{small}$\{s_1, \ldots, s_I\}$\end{small}; learning rate for inner update \begin{small}$\alpha$\end{small}; meta-learning rate \begin{small}$\beta$\end{small}; loss weighting factor \begin{small}$\lambda$\end{small}; length of sequence \begin{small}$|\mathcal{T}_c|$\end{small}}
    % \STATE Construct the market segment knowledge graph \begin{small}$\mathcal{G}$\end{small}
    Initialize all learnable parameters     \begin{small}$\Theta$\end{small}\;
    % \tcc{Meta-training Process}
    \While{not done}{
    Sample a batch of segments from \begin{small}$\{s_1, s_2, \ldots, s_I\}$\end{small}\;
    \For{each $s_i$}{
    Sample training set \begin{small}$\mathcal{D}_i^{tr}$\end{small} and testing set \begin{small}$\mathcal{D}_i^{te}$\end{small} from $s_i$\;
    Calculate the data-driven segment representation \begin{small}$\mathbf{q}_i^d$\end{small} by Eqn.~\eqref{eq:segment_repr}, reconstruction loss \begin{small}$\mathcal{L}_{rec}$\end{small} by Eqn.~\eqref{eq:rec}\;
    Calculate segment knowledge graph representation \begin{small}$\mathbf{q}_i^g$\end{small}and concatenate \begin{small}$\mathbf{q}_i^d$\end{small} and \begin{small}$\mathbf{q}_i^g$\end{small} as \begin{small}$\mathbf{q}_i$\end{small}\;
    Use \begin{small}$\mathbf{q}_i$\end{small} to customize the model parameter initialization by Eqn.~\eqref{eq:customize} and get \begin{small}$\theta_{0i}$\end{small}\;
    Optimize parameters starting from \begin{small}$\theta_{0i}$\end{small} as: \begin{small}$\theta_{i}=\theta_{0i}-\alpha\nabla_{\theta}\mathcal{L}(\mathcal{D}_i^{tr};\theta)$\end{small}\;
}
    Update \begin{small}$\Theta\leftarrow \Theta-\beta\nabla_{\Theta}\mathcal{L}_{joint}$\end{small} as Eqn.~\eqref{eq:joint_loss}\;
}
    % \tcc{Meta-testing Process}
    % \For{each target segment \begin{small}$s_t$\end{small}}{
    % Get training set \begin{small}$\mathcal{D}_t^{tr}$\end{small} and testing set \begin{small}$\mathcal{D}_t^{te}$\end{small}\;
    % Compute the data-driven representation and knowledge graph representation and then get $\mathbf{q}_t=\mathbf{q}_t^d\oplus \mathbf{q}_t^g$\;
    % Customize the learned model parameter initialization \begin{small}$\theta_{0}^{*}$\end{small} by using \begin{small}$\mathbf{q}_t$\end{small} and get the customized initialzations \begin{small}$\theta_{0t}^{*}$\end{small}\;
    % Update parameters as: \begin{small}$\theta_{t}=\theta^{*}_{0t}-\alpha\nabla_{\theta}\mathcal{L}(\mathcal{D}_t^{tr};\theta)$\end{small}\;
    % Evaluate and report the performance on \begin{small}$\mathcal{D}_{t}^{te}$\end{small} using optimized parameters \begin{small}$\theta_{t}$\end{small}}
\end{algorithm}

%% file: system.tex
\section{System Overview}
\label{sec:system}
\begin{figure}[htbp]
\centering
\includegraphics[width=0.35\textwidth]{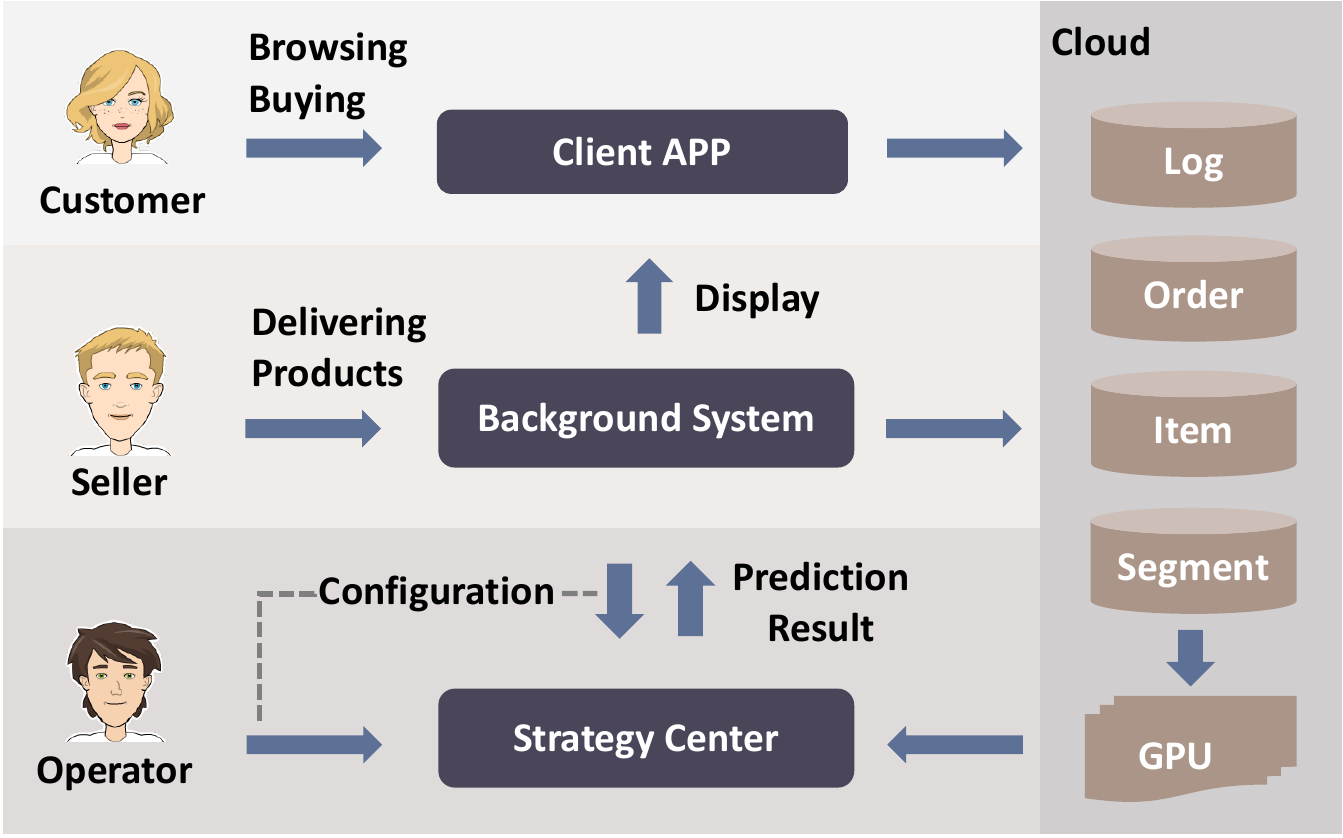}%
\caption{Online system overview}
\label{fig:system} 
\end{figure}
%back-office h
In this section, we introduce the system deployment pipeline of RMLDP in a real-world platform. We independently abstract the strategy center from the E-commerce background management system to implement our algorithm in business scenarios and maintain the low coupling and high cohesion in system design.

%基于模型的预测结果, 分配到不同的资源. 后台根据这种资源给商家分配display xxx, 最后在用户的手机端呈现
In Figure~\ref{fig:system}, we show the online system. In E-commerce platform, the operators first define the market segments based on a series of field configurations (e.g., the consumer groups served, price ranges of goods, brand collections, etc). The predicted segments are then selected, and the strategy center provides the demand prediction results using the proposed algorithm. Based on the prediction results, each segment's reasonable market flow resource is determined and sent to the background system. The background system assigns the specific display time and the display channel for each segment. Finally, the segment-specific information are displayed in the consumers' client App.

With the rapid development of cloud computing technology, behavioral logs (e.g., detailed product information, user's click, and purchase records on the client) are collected back to cloud storage in real-time. Based on this data, the MapReduce task deployed in the cloud extracts the required features and provides a steady feature stream. With the help of GPU, the model is efficiently trained. The whole system forms a complete closed loop.

%% file: experiment.tex
% !TEX root = main.tex
\section{Experiments}
\label{sec:experiments}
In this section, we conduct comprehensive experiments to evaluate our proposed RMLDP by answering the following research questions: (1) How is the overall prediction performance of RMLDP compared with state-of-the-art baselines? (2) How do various components we proposed (e.g., market segment knowledge graph) impact the model’s performance? (3) How is the online performance of RMLDP based on our proposed method and system?
\subsection{Experimental Setups}
In the experiment settings, we describe two real-world datasets and the compared baselines. The mean absolute percentage error (\textbf{MAPE}) are used to evaluate the performance.
\subsubsection{Dataset Description}
To evaluate our proposed method, we collect the data from two large-scale marketing scenarios in Taobao, the largest E-commerce platform in China~\cite{taobao}. We detail the descriptions as follows:
%The proposed model is tested on the datasets of two marketing scenarios in Taobao.
\begin{itemize}[leftmargin=*]
\item \textbf{Juhuasuan}~\cite{juhuasuan}: The first dataset is collected from Juhuasuan, one of the largest platform for group buying in China. There are more than 4000 segments over 800 days. We select three coarse-grained categories (electric appliances, clothing, and daily supplies) with nine fine-grained categories (large electric appliances, small electric appliances, digital electric appliances, women's clothing, men's clothing, sports, food, daily chemicals, daily sundry). 

\item \textbf{Tiantiantemai}~\cite{tiantiantemai}: Another dataset collected from Tiantiantemai, one of the largest platform for low-cost products in China. There are more than 6000 market segments over 189 days. The segments are selected from 3 coarse-grained categories (house hold, clothing, food) with 9 fine-grained categories (kitchenware, bedding, toiletries, womens' clothing, men's clothing, children's clothing, snacks, fresh, drink). 
%Similar to Juhuasuan, 70\% of segments are used for training, 10\% for validation, and 20\% are used for testing.
\end{itemize}
For both Juhuasuan and Tiantiantemai, we sort all segments by the number of purchasing records. We select top 70\% segments with more records for meta-training and the rest for meta-validation and meta-testing. For each fine-grained category in Juhuasuan and Tiantiantemai, in this experiment, the averaged performance (MAPE) over all segments with this category are reported.
\subsubsection{Hyperparameter Settings}
In Table~\ref{tab:hyperparameter}, we list all hyperparameters of Juhuasuan and Tiantiantemai.
\begin{table}[h]
\small
\caption{Hyperparameter Settings.}
\label{tab:hyperparameter}
\begin{center}
\begin{tabular}{l|cc}
\toprule
Hyperparameter & Juhuasuan & Tiantiantemai \\\midrule
batch size & 128 & 128 \\
feature dimension & 48 & 48\\
sequence length $|\mathcal{T}_c|$ & 30 & 30 \\
GRU embedding dimension & 128 & 128 \\
dimension of $\mathbf{q}_i^d$ & 32 & 32\\
dimension of $\mathbf{q}_i^g$  & 16 & 16\\
learning rate $\alpha$    & $10^{-4}$ & $10^{-4}$ \\
meta-learning rate $\beta$ & $10^{-3}$ & $10^{-3}$ \\
loss factor $\lambda$ & 0.5 & 0.5\\\bottomrule
\end{tabular}
\end{center}
\end{table}
\subsubsection{Baselines}
We compare our proposed method with the following four types of baselines: (1) \emph{Basic regression methods}: Linear Regression, Support vector regression (SVR); (2) \emph{Ensemble regression methods}: Random Forest, XGBoost~\cite{chen2016xgboost}; (3) \emph{Neural-network-based methods}: GRU, Dipole~\cite{ma2017dipole}, LSTNet~\cite{lai2018modeling}. For GRU and Dipole, we use the MPFN as backbone models and denote these two methods as GRU+MPFN and Dipole+MPFN, respectively. (4) \emph{Transfer methods}: Fine-tuning, MAML~\cite{finn2017model}, Meta-SGD~\cite{li2017meta}. In Fine-tuning, we use the same strategy as GRU+MPFN to learn the model parameters. Then, we finetune the learned parameters for each target segment. For all baselines, we use the same features as RMLDP. In basic regression, ensemble regression, and neural-network-based methods, the training dataset includes samples from all source segments and the training samples from target segments for a fair comparison.
\begin{table*}[h]
\small
\caption{Overall Performance of Juhuasuan.}
\begin{tabular}{l|c|c|c|c|c|c|c|c|c}
\toprule
\multirow{2}{*}{Model} & \multicolumn{3}{c|}{Electric Appliances} & \multicolumn{3}{c|}{Clothing} & \multicolumn{3}{c}{Daily Supplies} \\\cmidrule{2-10}
\multirow{2}{*}{} & Large  & Small & Digital & Women & Men & Sports & Food & Chemicals & Sundry \\
 \midrule
%  &  &  &  &  &  &  \\
%CNN &  &  &  &  &  &  \\
Linear Regression	&	42.19\%	&	43.26\%	&	41.43\%	&	46.32\%	&	47.13\%	&	47.94\%	&	41.02\%	&	42.24\%	&	43.97\%	\\
SVR	&	30.16\%	&	29.26\%	&	30.06\%	&	34.45\%	&	35.56\%	&	36.94\%	&	30.11\%	&	30.31\%	&	33.12\%	\\\midrule
Random Forest	&	$26.51$\%	&	$27.41$\%	&	$26.58$\%	&	$28.53$\%	&	$29.36$\%	&	$30.55$\%	&	$26.49$\%	&	$26.84$\%	&	$27.29$\%	\\
XGBoost~\cite{chen2016xgboost}	&	$25.38$\%	&	$26.81$\%	&	$25.19$\%	&	$27.49$\%	&	$28.37$\%	&	$28.76$\%	&	$26.93$\%	&	$26.53$\%	&	$26.07$\%	\\\midrule
GRU+MPFN~\cite{chung2014empirical}	&	$25.62$\%	&	$26.34$\%	&	$25.46$\%	&	$27.51$\%	&	$27.34$\%	&	$29.06$\%	&	$27.09$\%	&	$26.67$\%	&	$26.33$\%	\\
Dipole+MPFN~\cite{ma2017dipole}	&	$25.53$\%	&	$26.07$\%	&	$25.37$\%	&	$27.43$\%	&	$27.05$\%	&	$28.86$\%	&	$27.01$\%	&	$26.27$\%	&	$26.20$\%	\\
LSTNet~\cite{lai2018modeling}	&	25.98\%	&	26.66\%	& 26.31\%	& 27.48\%	&	27.56\%	& 29.13\%	&	28.93\%	&	26.94\%	& 26.89\%	\\\midrule
Finetune	&	24.29\%	&	26.20\%	&	24.13\%	&	27.12\%	&	26.98\%	&	28.23\%	&	26.54\%	&	26.01\%	&	26.01\%	\\
MAML~\cite{finn2017meta}	&	24.21\%	&	26.08\%	&	23.53\%	&	26.77\%	&	26.51\%	&	27.93\%	&	25.79\%	&	25.04\%	&	25.99\%	\\\midrule
\textbf{RMLDP$^{*}$}	&   \textbf{23.96\%}& \textbf{25.29\%}& \textbf{22.84\%}& \textbf{26.21\%}& \textbf{25.87\%	}& \textbf{26.98\%}& \textbf{24.25\%}& \textbf{24.38\%	}& \textbf{25.11\%}\\
\bottomrule
\end{tabular}
\\\vspace{0.1cm}
*: comparing with MAML, the results of RMLDP are significant according to Student's t-test at level 0.01.
\label{tab:results_juhuasuan}
\end{table*}

\begin{table*}[h]
\small
\caption{Overall Performance of Tiantiantemai.}
\begin{tabular}{l|c|c|c|c|c|c|c|c|c}
\toprule
\multirow{2}{*}{Model} & \multicolumn{3}{c|}{Household} & \multicolumn{3}{c|}{Clothing} & \multicolumn{3}{c}{Food} \\\cmidrule{2-10}
\multirow{2}{*}{} & Kitchenware & Bedding & Toiletries  & Women & Men & Children & Snacks & Fresh & Drink \\
 \midrule
Linear Regression	&	56.89\%	&	47.29\%	&	55.68\%	&	47.51\%	&	49.98\%	&	46.45\%	&	49.64\%	&	53.53\%	&	54.11\%	\\
SVR	&	37.16\%	&	36.26\%	&	38.25\%	&	37.81\%	&	38.94\%	&	38.55\%	&	38.29\%	&	38.19\%	&	39.25\%	\\\midrule
Random Forest	&	31.21\%	&	30.41\%	&	31.45\%	&	29.40\%	&	31.35\%	&	29.15\%	&	30.81\%	&	30.84\%	&	31.92\%	\\
XGBoost~\cite{chen2016xgboost}	&	31.16\%	&	30.81\%	&	30.86\%	&	29.49\%	&	31.43\%	&	29.46\%	&	30.13\%	&	30.49\%	&	31.07\%	\\\midrule
GRU+MPFN~\cite{chung2014empirical}		&	31.62\%	&	30.34\%	&	30.33\%	&	29.48\%	&	31.91\%	&	29.71\%	&	30.19\%	&	30.67\%	&	31.03\%	\\
Dipole+MPFN~\cite{ma2017dipole}		& 30.97\%	& 29.89\%	& 30.01\%	& 29.41\% 	&	30.69\%	& 30.12\%	& 30.14\%	&	30.33\%	&	30.66\%	\\
LSTNet~\cite{lai2018modeling}	&	31.49\%	&	30.76\%	&	30.12\% & 30.09\%	&	30.98\%	& 30.01\%	&	30.84\%	&	31.16\%	& 31.98\%	\\\midrule
Finetune	&	30.49\%	&	29.56\%	&	29.54\%	&	29.27\%	&	30.86\%	&	29.49\%	&	30.10\%	&	30.01\%	&	30.52\%	\\
MAML~\cite{finn2017meta}	&	29.55\%	&	29.08\%	&	29.41\%	&	29.06\%	&	30.07\%	&	29.31\%	&	29.97\%	&	29.84\%	&	30.01\%	\\\midrule
\textbf{RMLDP$^{*}$}	&   \textbf{	28.54\%	}& \textbf{	28.19\%	}& \textbf{	28.85\%	}& \textbf{	27.93\%	}& \textbf{	28.56\%	}& \textbf{	28.41\%	}& \textbf{	29.02\%	}& \textbf{	29.34\%	}& \textbf{	29.17\%	}\\
\bottomrule
\end{tabular}
\\\vspace{0.1cm}
*: comparing with MAML, the results of RMLDP are significant according to Student's t-test at level 0.01 
\label{tab:tiantiantemai}
\end{table*}
\subsection{Results}
\subsubsection{Overall Performance}
After implementing our proposed model and comparing it with other baselines, we report the results for Juhuasuan and Tiantiantemai in Table~\ref{tab:results_juhuasuan} and Table~\ref{tab:tiantiantemai}, respectively. For each fine-grained category, the averaged MAPE over segments in this category are reported. According to these results, we draw the following conclusions:
\begin{itemize}[leftmargin=*]
    \item All other types of baselines significantly outperform the basic regression methods (i.e., Linear regression, SVR). The reason is that it is non-trivial to capture complex non-linear temporal patterns through the basic regression methods.
    %\item Comparing with the neural-network-based methods (i.e., GRU-MPFN, Dipole-MPFN), ensemble regression methods achieve comparable performance. The results indicate that the ensemble models with 
    \item All transfer learning methods (i.e., MAML, Finetune and RMLDP) achieves better performance than other non-transfer methods. The results suggest that finetuning the learned knowledge from other segments can capture the task-specific information in the target segment and further benefit the performance. 
    \item In all cases, our RMLDP outperforms other baselines. In particular, RMLDP achieves better performance than MAML, which indicates the effectiveness of customizing model parameter initialization by leveraging the complex relations across market segments. Combining with the segment relations, the stability and diversity of transferred knowledge increases to the highest degrees.
\end{itemize}
\subsubsection{Ablation Study}
%RMLDP-qd. 去掉graph embedding RMLDP-gd. 去掉task embedding RMLDP-szn. 两条gru留第一条 RMLDP-local. 两条gru留第二条
We further perform comprehensive ablation studies to demonstrate the effectiveness of proposed components. We describe the ablation models as follows:
\begin{itemize}[leftmargin=*]
    \item \textbf{RMLDP-d}: In RMLDP-d, we remove the data-driven segment representation and only use the segment knowledge graph representation to modulate the model parameter initialization.
    \item \textbf{RMLDP-g}: In RMLDP-g, the segment knowledge graph is removed, and the data-driven market segment representation is the only signal for customizing model parameter initialization.
    \item \textbf{RMLDP-szn}: We only consider the local temporal trend in RMLDP-szn, i.e., the $\mathrm{GRU}^l$ is removed in the base learner MPFN.
    \item \textbf{RMLDP-local}: Contrary to RMLDP-szn, in RMLDP-local, we remove $\mathrm{GRU}^r$ in the base learner MPFN.
\end{itemize}

The results for Juhuasuan and Tiantiantemai are reported in Table~\ref{tab:ablation_juhuasuan} and Table~\ref{tab:ablation_tiantiantemai}, respectively. The performance of RMLDP is also reported for comparison. From these tables, we observe that:
\begin{itemize}[leftmargin=*]
    \item RMLDP performs better than both RMLDP-d and RMLDP-g, indicating the effectiveness and complementarity of segment knowledge graph representation and data-driven representation.
    \item Comparing with RMLDP-d, RMLDP-g achieves better performance. The potential reason is that the data-driven market segment representations, which are learned from training data of each segment, capture the segment-specific temporal patterns, and provide more effective information.
    \item RMLDP significantly outperforms RMLDP-szn and RMLDP-local, indicating that both local and seasonal temporal patterns contribute to the model performance. The seasonal temporal patterns provide the basic estimation for the segment demand, and the local temporal patterns further provide the calibration by using the most recent records.
\end{itemize}
\begin{table*}[h]
\small
\caption{Ablation studies of Juhuasuan.}
\begin{tabular}{l|c|c|c|c|c|c|c|c|c}
\toprule
\multirow{2}{*}{Model} & \multicolumn{3}{c|}{Electric Appliances} & \multicolumn{3}{c|}{Clothing} & \multicolumn{3}{c}{Daily Supplies} \\\cmidrule{2-10}
\multirow{2}{*}{} & Large  & Small & Digital & Women & Men & Sports & Food & Chemicals & Sundry \\
 \midrule
RMLDP-d	&	24.18\%	&	26.03\%	&	23.41\%	&	26.45\%	&	26.47\%	&	27.84\%	&	25.63\%	&	25.01\%	&	25.97\%	\\
RMLDP-g	&	24.09\%	&	25.84\%	&	22.97\%	&	26.28\%	&	26.01\%	&	27.53\%	&	24.81\%	&	24.77\%	&	25.44\%	\\

RMLDP-szn	&	28.01\%	&	27.97\%	&	26.67\%	&	33.29\%	&	34.72\%	&	35.88\%	&	29.31\%	&	29.34\%	&	31.51\%	\\
RMLDP-local	&	30.42\%	&	31.38\%	&	31.23\%	&	34.98\%	&	35.87\%	&	37.32\%	&	33.09\%	&	31.09\%	&	33.49\%	\\\midrule
\textbf{RMLDP}	&   \textbf{	23.96\%	}& \textbf{	25.29\%	}& \textbf{	22.84\%	}& \textbf{	26.21\%	}& \textbf{	25.87\%	}& \textbf{	26.98\%	}& \textbf{	24.25\%	}& \textbf{	24.38\%	}& \textbf{	25.11\%	}\\
\bottomrule
\end{tabular}
\label{tab:ablation_juhuasuan}
\end{table*}

\begin{table*}[h]
\small
\caption{Ablation studies of Tiantiantemai.}
\begin{tabular}{l|c|c|c|c|c|c|c|c|c}
\toprule
\multirow{2}{*}{Model} & \multicolumn{3}{c|}{Household} & \multicolumn{3}{c|}{Clothing} & \multicolumn{3}{c}{Food} \\\cmidrule{2-10}
\multirow{2}{*}{} & Kitchenware & Bedding & Toiletries  & Women & Men & Children & Snacks & Fresh & Drink \\
 \midrule
RMLDP-d	&	29.35\%	&	29.07\%	&	29.36\%	&	29.01\%	&	29.96\%	&	28.53\%	&	29.89\%	&	29.78\%	&	29.86\%	\\
RMLDP-g	&	28.91\%	&	28.65\%	&	29.04\%	&	28.09\%	&	29.33\%	&	29.08\%	&	29.15\%	&	29.41\%	&	29.31\%	\\
RMLDP-szn	&	33.83\%	&	33.54\%	&	34.99\%	&	37.36\%	&	36.78\%	&	37.98\%	&	37.54\%	&	36.54\%	&	38.27\%	\\
RMLDP-local	&	41.96\%	&	37.29\%	&	42.53\%	&	40.15\%	&	42.39\%	&	40.49\%	&	41.04\%	&	42.93\%	&	41.45\%	\\\midrule
\textbf{RMLDP}	&   \textbf{	28.54\%	}& \textbf{	28.19\%	}& \textbf{	28.85\%	}& \textbf{	27.93\%	}& \textbf{	28.79\%	}& \textbf{	28.41\%	}& \textbf{	29.02\%	}& \textbf{	29.34\%	}& \textbf{	29.17\%	}\\
\bottomrule
\end{tabular}
\label{tab:ablation_tiantiantemai}
\end{table*}
%1. 去掉graph embedding 2. 去掉task embedding 3. 去掉reconstruction loss 4. 两条gru留第一条 5. 两条gru留第二条
\subsubsection{Effect of Sequence Length}
\begin{figure}[h]
	\centering
 	\includegraphics[width=0.42\textwidth]{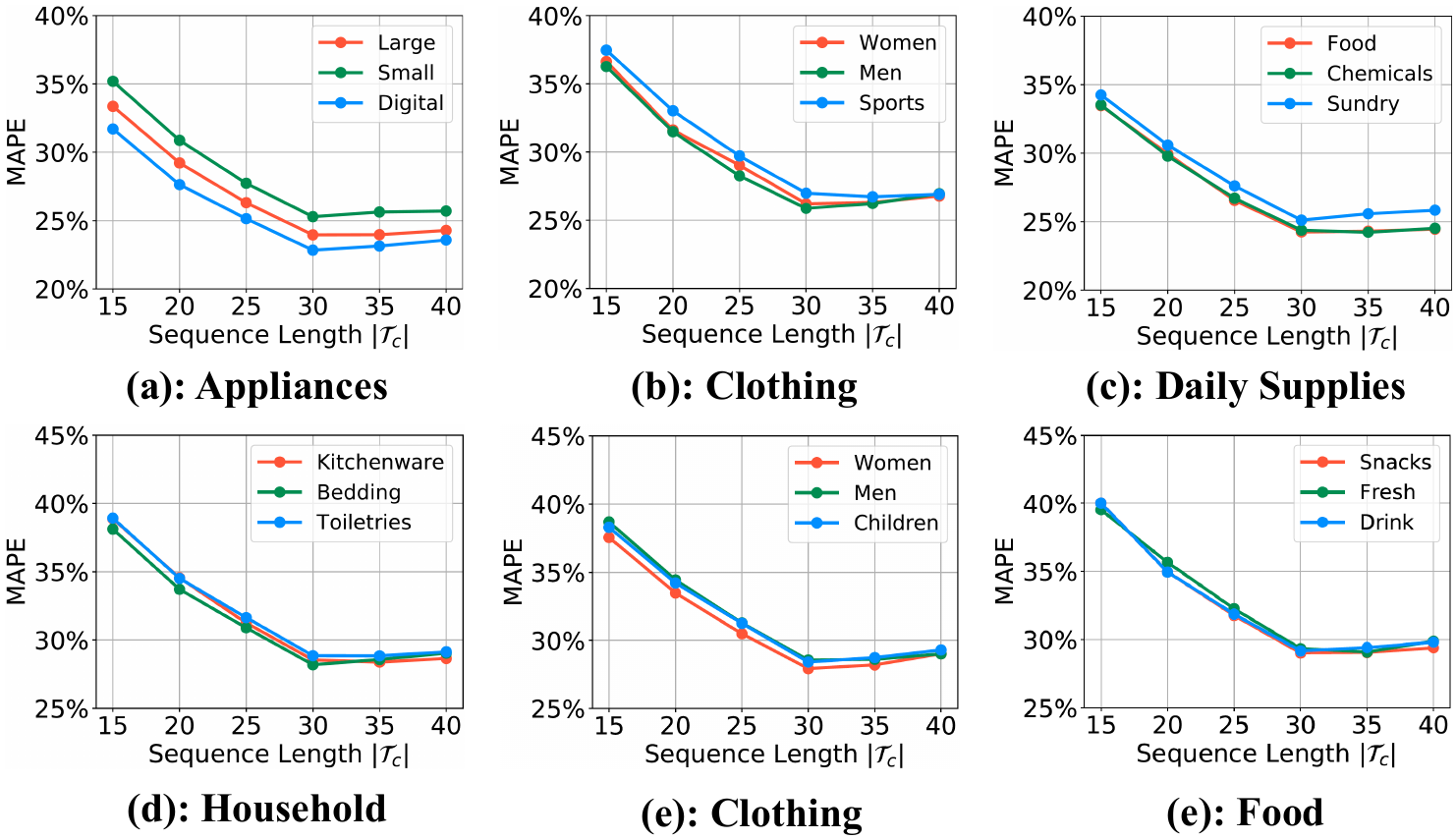}
	\caption{Prediction performance on each coarse-category from v.s. the sequence length $|\mathcal{T}_c|$. (a), (b), (c): results on Juhuasuan; (d), (e), (f): results on Tiantiantemai.} 
	\label{fig:seq_len}
\end{figure}

In this section, we analyze the effect of sequence length (i.e., the value of $|\mathcal{T}_c|$). We change the sequence length from 15 to 40, and the results for each coarse-category of two datasets are shown in Figure~\ref{fig:seq_len}. We can see that the MAPE decreases at the beginning and then keeps stable/slightly increases. The reason is that too short sequence may not provide enough information for accurate prediction. When the length of sequence increases, the covered information gradually becomes saturated, and the results keep stable.
%1. 随着样本数据量的关系 2. 序列长度的影响
\subsubsection{Analysis of Segment Representation} 
\begin{figure}[h]
	\centering
 	\includegraphics[width=0.35\textwidth]{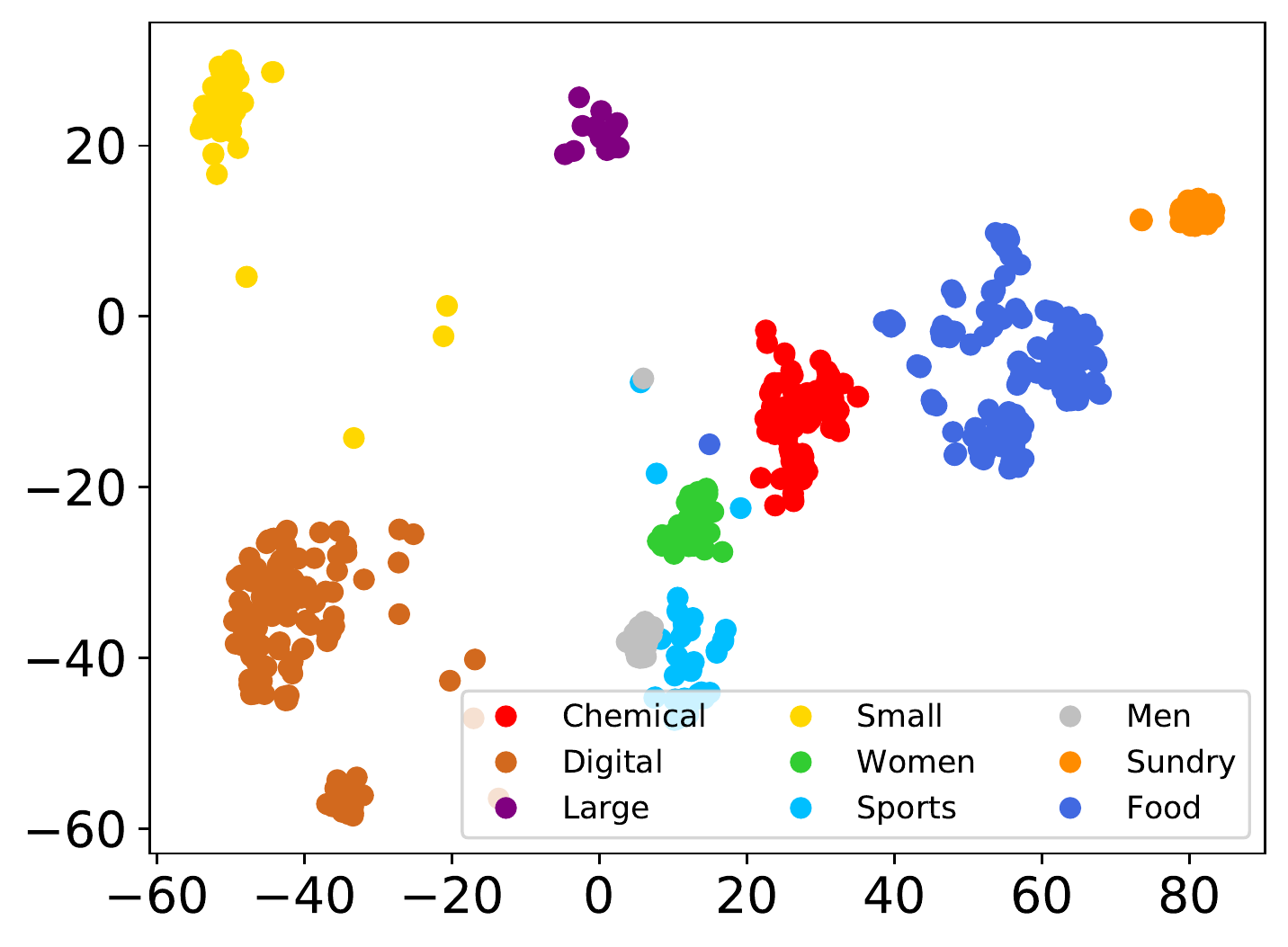}
	\caption{Visualization of learned segment representation from 694 meta-testing segments.} 
	\label{fig:tsne}
\end{figure}

We further analyze the segment representation $\mathbf{q}_{i}$ discussed Section~\ref{sec:representation}, where 694 meta-testing segments in Juhuasuan are used. The results are shown in Figure~\ref{fig:tsne}. We observe that the segment representations are capable of well-distinguishing different categories of segments and further provide qualitative evidence for the effectiveness of RMLDP.
%and further verify the better performance of RMLDP by considering the segment information.
%1. modulate 之前的task embedding 2.最后学到的embedding
\subsubsection{Online Experiment}
To further evaluate the proposed model, we design the online experiments in Taobao mobile App. We conduct a bucket testing (i.e, A/B testing) in Tiantiantemai to test the consumers' response to our RMLDP and baseline. For each segment, the higher demand prediction value it gets, the more opportunities of display it gains.

Without using prediction model, operators usually leverage the averaged demand value from the same period of the last year and the nearest month to predict the future demand. In offline evaluation, the MAPE for this statistical method is more than 0.8. We regard this statistical method as our baseline, and calculate five core indicators: Page View (PV), Unique Visitor (UV), total number of segments with orders (\#Seg), total number of products with orders (\#Item), weekly orders (Ord). The results are reported in Table~\ref{tab:online1}. Except for the supply for different market segments, both buckets have the same personalization strategy of recommendation system. Comparing with the statistical method, our model achieves better performance under the similar volumes of PageView and Unique Visitor.

\begin{table}[h]
\small
\caption{The results of different prediction strategies}
\label{tab:online1}
%\begin{center}
\begin{tabular}{l|ccccc}
\toprule
Bucket &  PV & UV & \#Seg & \#Item  & Ord \\\midrule
Stat. Method & 1.93M & 0.64M  & 1334 & 3708   & 91980 \\
\textbf{RMLDP} & 1.92M & 0.64M & \textbf{1458} & \textbf{3809}  & \textbf{94780} \\\bottomrule
\end{tabular}
%\end{center}
\end{table}
%\caption{The results of different segment display strategies.}

%% file: conclusion.tex
% !TEX root = main.tex
\section{Conclusion}
In this paper, we propose a novel relation-aware meta-learning framework, RMLDP, for market segment demand prediction with limited data by transferring knowledge from data-sufficient segments. Our proposed method incorporates the base demand prediction model (i.e., multi-pattern fusion network) into a meta-learning paradigm. The model parameter initialization learned from source segments can be easily adapted to each target segment. Additionally, the segment relations are learned and tailored to customize the transferable model initialization. Extensive experiments on two large-scale E-commerce datasets verify the effectiveness of RMLDP. RMLDP is 
further deployed in the real-wold platform with the positive bucket testing results.
\label{sec:conclusion}